\documentclass[10pt,a4paper,twocolumn]{article}
\usepackage[a4paper,top=0.7in,bottom=0.8in,left=0.65in,right=0.65in]{geometry}
\usepackage{graphicx}
\usepackage{booktabs}
\usepackage{wrapfig}
\usepackage{float}
\usepackage{enumitem}
\usepackage{diagbox}
\usepackage{xspace}
\usepackage{makecell}
\usepackage{color}
\usepackage{xcolor}
\usepackage{pifont}
\usepackage{multirow}
\usepackage{xspace}
\usepackage{wrapfig,lipsum}
\usepackage{diagbox}
\usepackage{caption}
\usepackage{cuted}  
\usepackage{tabularx}
\usepackage{hyperref}
\def\ourina{{\textit{iterNA}}\xspace}
\def\oursuffix{{\textit{suffL}}\xspace}
\def\ourmethod{{\textit{TFinv}}\xspace}

\newcommand{\cmark}{\ding{51}\xspace}%
\newcommand{\xmark}{\ding{55}\xspace}%
\usepackage{colortbl}
\newcommand{\quotes}[1]{``#1''}

\newcommand{\model}{\epsilon_\theta}

\newcommand{\conditioner}{\psi_\xi}

\newcommand{\expec}{\mathbb{E}}
\newcommand{\encoder}{\mathcal{E}}
\newcommand{\decoder}{\mathcal{D}}

\newcommand{\textprompt}{\mathcal{P}}
\newcommand{\textembedding}{\mathcal{C}_{text}}

\newcommand{\mlp}{\mathit{l}}
\newcommand{\sdturbo}{\mathcal{G}}

\newcommand{\inputimage}{\mathcal{I}}

\usepackage{amssymb}
\usepackage{amsmath}

\usepackage[numbers,sort&compress]{natbib}

\title{\textbf{Training-Free Image Inversion for One-Step Diffusion Models}}

\author{
Tao Wu$^{1}$,
Senmao Li$^{2}$,
Yaxing Wang$^{3}$,
Shiqi Yang$^{4}$,
Kai Wang$^{5,6}$,
Joost van de Weijer$^{1}$
\\[4pt]
\small $^{1}$Computer Vision Center, Universitat Autònoma de Barcelona, Spain\\
\small $^{2}$Mohamed bin Zayed University of Artificial Intelligence (MBZUAI), Abu Dhabi, United Arab Emirates\\
\small $^{3}$Jilin University, China\\
\small $^{4}$Independent Researcher, Tokyo, Japan\\
\small $^{5}$City University of Hong Kong (Dongguan), China \\
\small $^{6}$City University of Hong Kong, Hong Kong Special Administrative Region of China
}

\date{}

\begin{document}

\maketitle

\begin{abstract}
In this work, we introduce a novel \textit{training-free inversion}
(\ourmethod) framework for \textit{one-step diffusion models},
addressing key challenges in real image inversion and editing.
We first identify two critical factors hampering real-image inversion
and editing: (1) \textit{Initial Latent Editability}, which is related
to the distance between the initial noise and the ideal Gaussian
distribution, and (2) \textit{Caption Gap}, which means the alignment
between text captions and image representations. Both factors influence
inversion efficiency and the editability of one-step diffusion models.
Then, we propose two novel techniques: \textit{iterative noise alignment}
(\ourina), which minimizes the distribution gap to align with the normal
Gaussian distribution, and \textit{suffix learning} (\oursuffix), which
enhances text-to-image caption alignment by introducing learned suffix
prompt tokens. These techniques enable precise inversion of input images
into their initial noise representations and facilitate image editing.

Furthermore, we propose a mask-based editing technique for localized
edits while preserving background integrity. Comprehensive experiments
on the \textit{PIE-Bench} dataset validate that our method \ourmethod
not only achieves state-of-the-art performance in
\textit{one-step diffusion editing}, but also significantly outperforms
existing multistep approaches in efficiency.
The code is available at
\url{https://github.com/tttao-uwu/TFinv.git}.
\end{abstract}






\section{Introduction}

Recent T2I diffusion models~\cite{Rombach_2022_CVPR_stablediffusion,luo2023LCM} excel in generating high-quality, text-aligned images but rely on computationally demanding multistep sampling~\cite{song2021ddim}, gradually refining noise into coherent visuals. 
To address the limitation, recent research has explored methods to significantly reduce the number of sampling steps. 
Techniques employing \textit{few-step} sampling or even \textit{one-step} generation~\cite{luo2023LCM,sauer2023adversarial} through distillation have emerged as promising solutions. 
These methods maintain high visual fidelity while substantially improving efficiency, enabling more efficient manipulation for downstream applications, such as image editing~\cite{deutch2024turboedit,wu2025turboedit_adobe,garibi2024renoise,huberman2024ddpm_friend,wang2026point2pix}.

For text-guided image editing~\cite{hertz2022prompt,tumanyan2023plug,miyake2023NPI}, existing approaches often utilize the DDIM~\cite{song2021ddim} inversion process to approximate the initial noise corresponding to a source image. This enables accurate reconstruction of the source image and facilitates content modification aligned with the guided text, while preserving non-edited regions. Starting from the inverted noise, techniques such as attention manipulation and hijacking~\cite{hertz2022prompt,kai2023DPL} are applied iteratively at each denoising step to introduce edits while maintaining the integrity of background elements.
While effective, these methods rely on two computationally intensive multistep processes—\textit{inversion and editing}—making them unsuitable with \textit{few-step diffusion models}. 
Recent efforts~\cite{deutch2024turboedit,wu2025turboedit_adobe} have sought to improve efficiency by adopting few-step diffusion models such as SD-Turbo~\cite{sauer2023adversarial}, which reduce the number of sampling steps required for both inversion and editing. These approaches further incorporate additional guidance mechanisms to achieve disentangled and precise editing based on text prompts.
For instance, TurboEdit~\cite{deutch2024turboedit} focuses on few-step diffusion-based editing by shifting the noise schedule with timestep offsets, but it performs poorly with one-step diffusion models. 
On the other hand, SwiftEdit~\cite{nguyen2024swiftedit} achieves one-step diffusion-based inversion and editing via training an inversion network, but this method requires resource-intensive training on large datasets for each specific one-step diffusion model and that significantly limits its applicability in diverse scenarios.

\begin{figure*}[t]
\begin{minipage}[t]{0.489\linewidth}
    \centering
    \includegraphics[width=\linewidth]{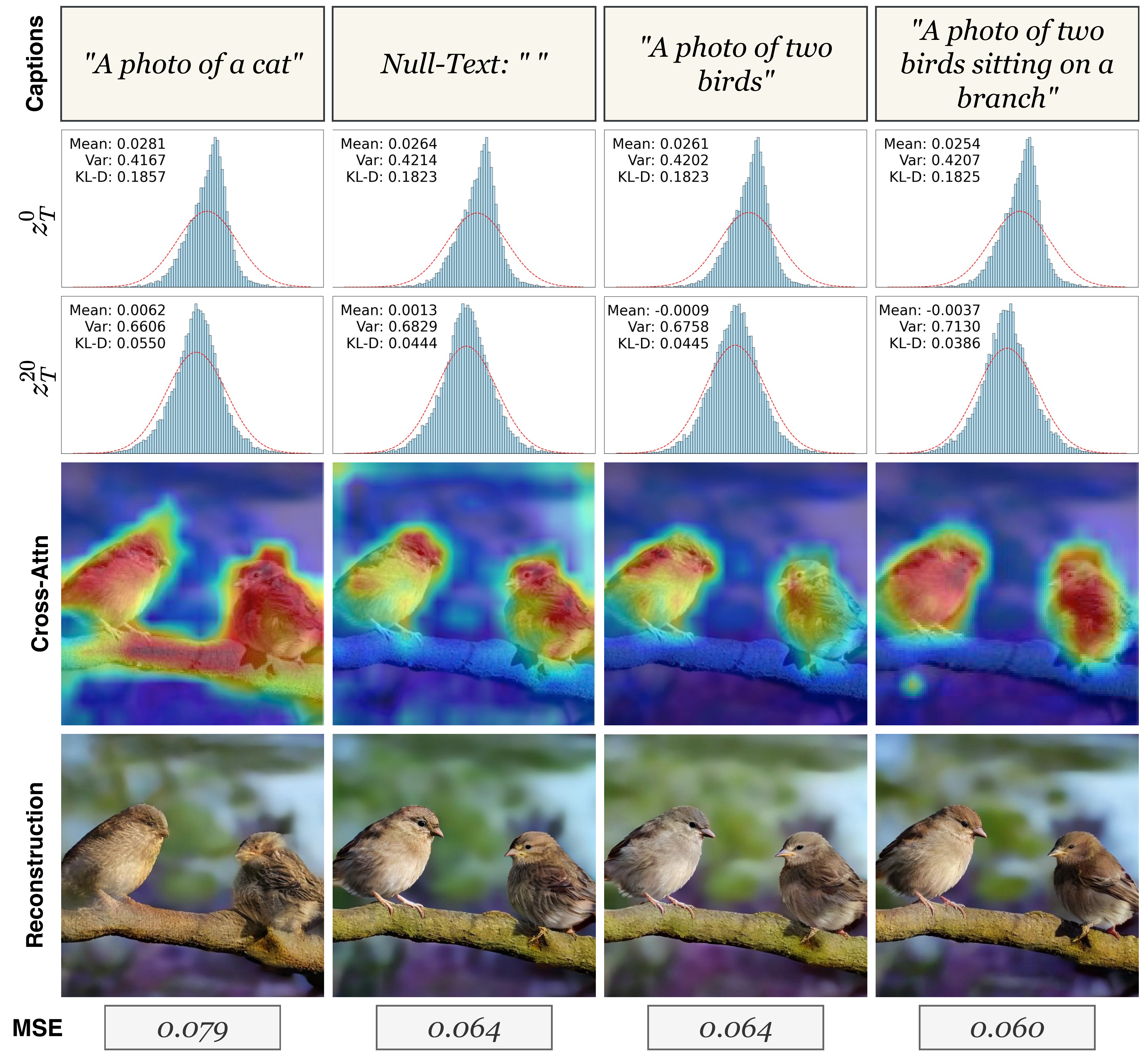}
    \vspace{-1mm}
    \caption{
To explore how the \textit{caption gap} influences the one-step diffusion inversion and editing, we analyze the impact of various caption conditions on the inversion process, including incorrect captions, null-text captions, human-written general captions, and captions generated by LLaVa~\cite{liu2024visual_llava,liu2024improved_llava}.
}

    \label{fig:anal_noise}
\end{minipage}
\hfill
\begin{minipage}[t]{0.489\linewidth}
    \centering
    \includegraphics[width=\linewidth]{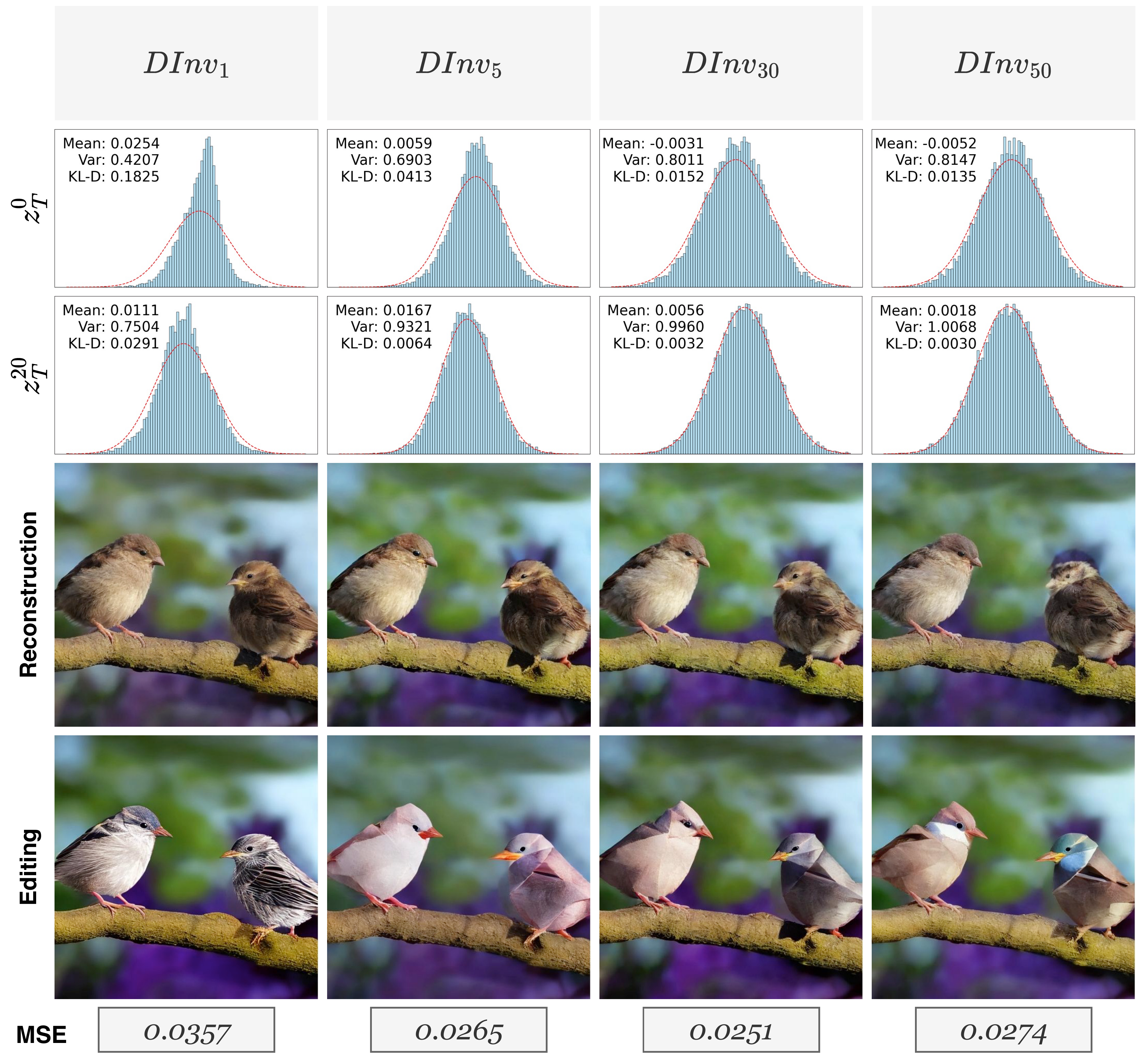}
    \vspace{-1mm}
    \caption{  
DDIM inversion ($DInv_{\tau}$) presents a trade-off between editability and reconstruction while with the same input caption. 
A smaller number $\tau$ of DDIM steps makes it more challenging to update the latent variable $z_T$ into a normal Gaussian distribution, yet it maintains more \textit{structural layout} information for image reconstruction in the initial latent space. 
}
    \label{fig:anal_steps}
\end{minipage}
\vspace{3mm}
\end{figure*}

Image editing is inherently iterative, where a good user experience crucially depends on rapid feedback across multiple refinement cycles. This creates a strong demand for efficient editing with one-step diffusion models. However, a fundamental bottleneck remains: inverting one-step diffusion models. Existing inversion techniques~\cite{song2021ddim,mokady2022null,li2023stylediffusion,huberman2024ddpm_friend} fail to reliably reconstruct input images or support effective editing. These shortcomings arise primarily from diffusion gaps~\cite{zhang2024real_bias} and signal leakage~\cite{everaert2024_signal_leak_bias}, which disrupt the inversion process.
These shortcomings arise primarily from diffusion gaps~\cite{zhang2024real_bias} and signal leakage~\cite{everaert2024_signal_leak_bias}, which disrupt the inversion process.
In this work, we start with an analysis of reconstruction challenges across diverse caption sources, which we term as the \textit{caption gap}. Our findings in Fig.~\ref{fig:anal_noise} reveal that inaccurate or misaligned captions lead to more biased initial noise distributions, making it harder to reconstruct the input image. This highlights the critical role of highly aligned image captions in the inversion process.
Additionally, we examine the initialization of the noisy latent using DDIM inversion with varying diffusion steps to reveal the \textit{initial latent editability}, as shown in Fig.~\ref{fig:anal_steps}. While increasing the number of inversion steps brings the initial noise distribution closer to the Gaussian distribution, it also significantly complicates the reconstruction, underscoring the trade-off between distribution alignment and reconstruction difficulty.

Building on our observations, we present the first \textit{Training-Free Inversion} (\ourmethod) framework for \textit{one-step diffusion models}, introducing two key techniques for real image inversion: \textit{Iterative Noise Alignment} (\ourina) and \textit{Suffix Learning} (\oursuffix). 
More specifically, \ourina ensures that the initial noise aligns closely with the ideal distribution, represented as a standard Gaussian distribution, by optimizing a Kullback–Leibler divergence objective. In addition, \oursuffix further refines the reconstruction by learning a set of suffix prompts of the captions that complement the reconstruction details, enhancing fidelity and preserving intricate features.
These innovations enable the inversion of input images into their initial noise representations, allowing accurate image reconstruction and seamless editing.  
Once the inversion process is complete, editing is performed by simply modifying the source prompts to the target prompts. 
Furthermore, we propose an efficient \textit{mask-based} editing technique that generates masks directly from the trained inversion network and guidance prompts, enabling effective blending and control of edit strength while preserving background elements.

To the best of our knowledge, this work is the first to propose training-free image editing leveraging one-step diffusion models. Our approach not only achieves significantly faster performance compared to existing few-step editing methods but also delivers competitive results in terms of editing quality, demonstrating its efficacy and practicality. 
In summary, our main contribution include:

\begin{itemize}[leftmargin=*]
    \item We conduct an examination of the inversion process by analyzing various input captions and diverse inversion steps. This comprehensive evaluation provides an in-depth understanding of two key factors: the highly aligned captions for accurate inversion and the inherent editing-reconstruction trade-off problem associated with varying inversion steps.
    
    \item Based on these observations, we introduce the first training-free image inversion and editing framework, \ourmethod, specifically designed for one-step diffusion models. Our approach leverages two novel real-image inversion techniques of our proposal: the iterative noise alignment (\ourina) and suffix learning (\oursuffix).
    
    \item Furthermore, we propose a cross-attention-based technique for mask-guided editing, enabling flexible and precise control over the editing strength. This approach integrates a mask generation process, ensuring that edits are seamlessly applied to the target regions while preserving critical background details.
    
    \item Through extensive experiments on the PIE-Bench dataset~\cite{ju2023direct}, our method, \ourmethod, achieves state-of-the-art performance in text-guided image editing compared to existing approaches utilizing one-step diffusion models.
\end{itemize}

\section{Related works}

\subsection{Text-to-Image Diffusion Models}

Text-to-image diffusion models~\cite{Rombach_2022_CVPR_stablediffusion,saharia2022imagen} have achieved remarkable performance in image generation. These models enable users to input a text prompt and generate images of unprecedented quality, thanks to their powerful image synthesis capabilities. However, the iterative denoising process required in diffusion models makes the \textit{inference process} time-consuming. To address this challenge, distillation techniques~\cite{yin2024onestep_dmd,luo2024diff_instruct} have been proposed to speed up the multi-step diffusion models into few-step ones~\cite{nguyen2023swiftbrush,sauer2023adversarial}.
In step with advances in generative AI, rapid progress has been made in text-to-image (T2I) generation. In particular, T2I diffusion models~\cite{Rombach_2022_CVPR_stablediffusion, saharia2022imagen} emerged as more efficient models surpassing GANs~\cite{karras2021alias, pernuvs2025fice}, VAEs~\cite{kingma2013vae}, autoregressive~\cite{esser2021taming,razavi2019generating,van2016pixel} and flow-based~\cite{dinh2016density} models in T2I generation. 

\subsection{Few-step Diffusion Models}
Several attempts have been made to accelerate the sampling process of diffusion models by introducing additional training beyond the base diffusion model.
One of the most representative methods is the \textit{Consistency Models}~\cite{song2023consistency,luo2023latent}, which leverage the inherent properties of ODE samplers within diffusion models to train a new model by minimizing the difference between points along identical trajectories. 
More recently, \textit{distillation} techniques~\cite{luo2024diff_instruct} have been applied to diffusion models, allowing faster training of student models by pre-trained teachers~\cite{sauer2023adversarial,nguyen2023swiftbrush}. 
For example, SD-Turbo~\cite{sauer2023adversarial} introduces a discriminator combined with a score distillation loss to improve performance.  
SwiftBrush~\cite{nguyen2023swiftbrush} adapts variational score distillation, initially developed for text-to-3D, into the text-to-image domain. 
In this work, we mainly build \ourmethod on the SD-Turbo~\cite{sauer2023adversarial} to accomplish the training-free real image inversion and editing.

\subsection{Text-guided Image editing} 

Text-guided image editing methods~\cite{tang2023iterinv,tang2024locinv,lin2024text_driven_edit} of
recent researches~\cite{huang2023kv} in this topic adopt the large pretrained T2I models~\cite{chang2023muse,ramesh2022dalle2} for controllable image editing.
Among them, Imagic~\cite{kawar2022imagic} and P2P~\cite{hertz2022prompt} attempt structure-preserving editing via Stable Diffusion (SD) models. Imagic~\cite{kawar2022imagic} requires fine-tuning the entire model for each image. P2P~\cite{hertz2022prompt} has no need to fine-tune the model and retrains the image structure by assigning \textit{cross-attention} maps from the original image to the edited one in the corresponding text token.

\subsection{Diffusion Inversion} 
As an important solution for text-guided image editing, diffusion-based inversion~\cite{miyake2023NPI,tang2024locinv,direct_inversion_2023} can be performed naively by optimizing the latent representation, which shows potential in editing tasks by deterministically calculating and encoding context information into a latent space and then reconstructing the original image using this latent representation. 
DDIM inversion~\cite{dhariwal2021diffusion} shows that a given real image can be reconstructed by DDIM sampling~\cite{song2021ddim}. DDIM provides a good starting point to synthesize a given real image.  
However, DDIM is found lacking for text-guided diffusion models when classifier-free guidance (CFG) is applied, which is necessary for meaningful editing.
Leveraging optimization on null-text embedding, NTI~\cite{mokady2022null} further improved the image reconstruction quality when CFG is applied and retained the rich text-guided editing capabilities of the SD model~\cite{Rombach_2022_CVPR_stablediffusion}. 
To enable text-guided editing on one-step diffusion models, prior approaches such as TurboEdit~\cite{deutch2024turboedit} adapt the training-free DDPM-Inversion~\cite{huberman2024ddpm_friend} for 4-step diffusion models. In contrast, methods like TurboEdit~\cite{wu2025turboedit_adobe} and SwiftEdit~\cite{nguyen2024swiftedit} employ resource-intensive training of specifically designed networks to predict the inverted noises.

In this paper, we focus on achieving effective and efficient prompt-based editing performance on one-step diffusion models—a challenge that existing methods have yet to address.
\begin{figure*}[t]
\centering
  \includegraphics[width=0.99\textwidth]{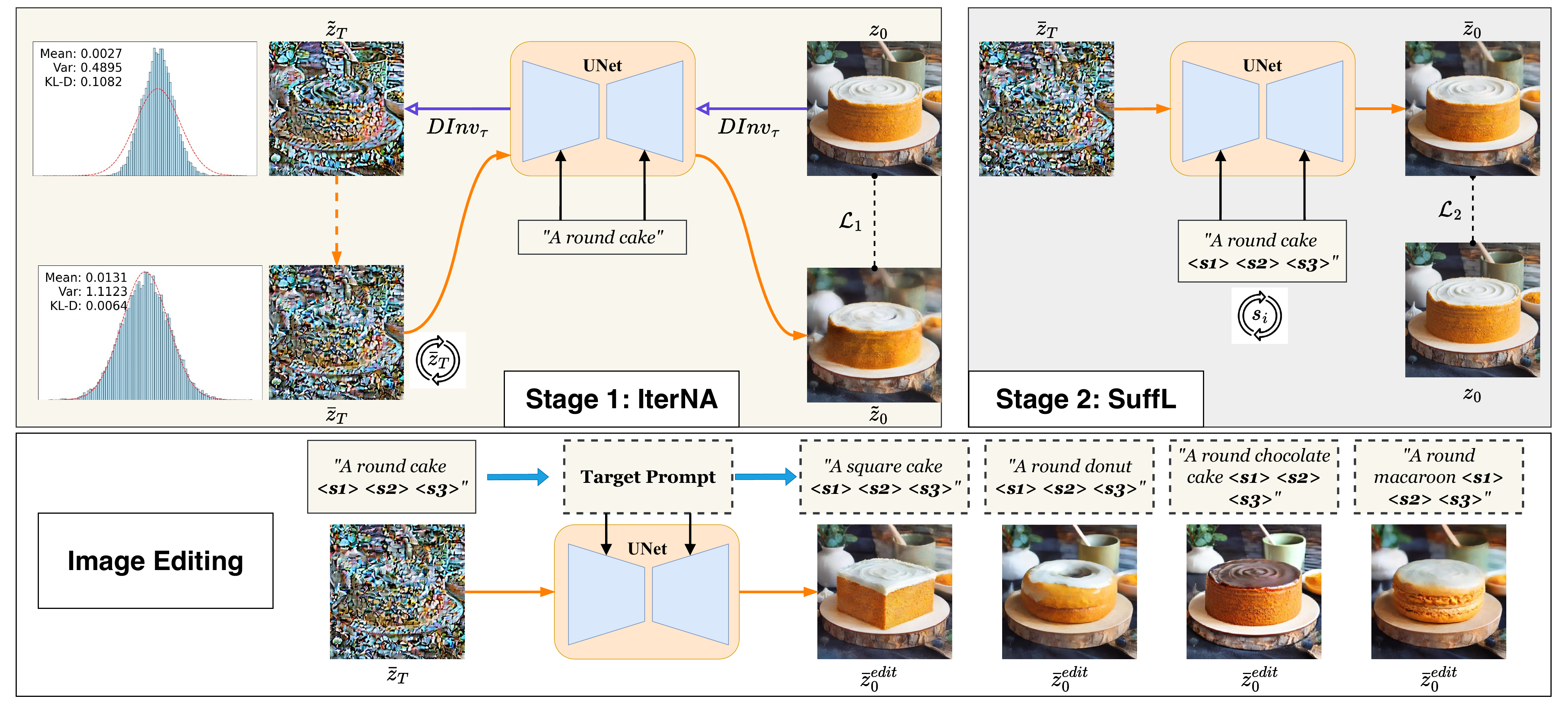}
  \caption{Our method \ourmethod is composed of two stages: the iterative noise alignment (\ourina) and suffix learning (\oursuffix). The \ourina stage focuses on refining the noisy latent $\tilde{z}_T$, ensuring it aligns closely with the ideal normal Gaussian distribution while maintaining the ability to reconstruct the real image $z_0$. 
  The \oursuffix stage further enhances reconstruction quality by updating the textual conditioning mechanism, which is achieved by introducing additional learnable suffix tokens appending to the textual prompt. These tokens contribute to improving reconstruction without compromising the flexibility of the editing process. 
  Finally, real image editing is performed effortlessly by modifying the conditional textual prompts. This process can be further augmented by employing our cross-attention-based mask mechanism depicted in Sec.~\ref{sec:image_editing}, which preserves critical background details while focusing edits on specified regions.}
  \vspace{-2mm}
  \label{fig:method}
\end{figure*}

\section{Method}

In this section, we present \ourmethod, a novel training-free framework for image editing within one-step diffusion models. In section~\ref{preliminaries}, we provide an introduction to the fundamental concepts of the diffusion model. Subsequently, we expound upon the design of our framework in Section~\ref{tfi}, which includes two stages, namely \ourina and \oursuffix. Finally, we delve into the editing process in Section~\ref{tfi_edit}.

\subsection{Preliminaries} \label{preliminaries}

\subsubsection{Latent Diffusion Models}
Latent Diffusion Model (LDM)~\cite{Rombach_2022_CVPR_stablediffusion} is the most widely applied T2I diffusion models and the distillation teacher model for current few-step diffusion models~\cite{sauer2023adversarial,luo2023LCM}, which consists of two main components: (i) An auto-encoder --- that transforms an image $\inputimage$ into a latent code $z_0=\encoder(\inputimage)$ while the decoder reconstructs the latent code back to the original image such that $\decoder(\encoder(\inputimage)) \approx \inputimage$; and (ii) Diffusion model ---
can be conditioned using class labels, segmentation masks, or textual input. Let $\conditioner(\textprompt)$ represent the textual conditioning mechanism that converts a condition $\textprompt$ into a conditional vector for LDMs. The LDM model is refined using the noise reconstruction loss:
\begin{equation}
    \mathcal{L}_{LDM} = \expec_{z_0 \sim \encoder(x), \epsilon \sim \mathcal{N}(0, 1)} \Vert \epsilon - \model(z_{t},t, \conditioner(\textprompt)) \Vert_{2}^{2} 
    \label{eq:loss}
\end{equation}
The backbone $\model$ is a conditional UNet which predicts the added noise. In particular, text-guided diffusion models aim to generate an image from the random noise $z_T$ and a conditional prompt $\textprompt$. 
We further itemize textual condition as $\textembedding=\conditioner(\textprompt)$. 
where $\conditioner$ is the CLIP text encoder~\citep{radford2021clip}.
The cross-attention map is derived from $\model(z_t,t,\textembedding)$. 
They are computed from the deep features of noisy image $\zeta(z_t)$ which are projected to a query matrix $Q_t=\mlp_Q (\zeta(z_t))$, and the textual embedding which is projected to a key matrix $K = \mlp_K (\textembedding)$. Then the attention map is computed according to: 
    $\mathcal{A}_t=softmax(Q_t \cdot K^T / \sqrt{d})$
where $d$ is the latent dimension, and the cell $[\mathcal{A}_t]_{ij}$ defines the weight of the $j$-th token on the latent pixel $i$.
After predicting the noise, diffusion schedulers~\cite{song2021ddim,lu2022dpm} are used to predict the latent $z_{t-1}$. As an example with the deterministeic DDIM scheduler~\cite{song2020denoising}, the formula is: 
\begin{equation}\label{eq:ddim_sampling}
\resizebox{0.9\linewidth}{!}{$
\boldsymbol{z_{t-1}} = \sqrt{\frac{\alpha_{t-1}}{\alpha_t}}\boldsymbol{z_{t}} + \sqrt{\alpha_{t-1}}\left(\sqrt{\frac{1}{\alpha_{t-1}}-1}-\sqrt{\frac{1}{\alpha_t}-1}\right) \cdot \model(z_t,t,\textembedding)
$}
\end{equation}
where $\alpha_t$ is a predefined scalar function. 

To accelerate the diffusion sampling process, various methods distill the original sampling steps $T_{teacher}=[1,T]$ of the teacher model into few student anchor steps $T_{student}={\upsilon_1,\upsilon_2,...,\upsilon_n}$ where $n={1,2,4}$ in most cases.
More specifically, \textit{one-step} diffusion model $\sdturbo$ aims to transform a noise input $z_T \sim \mathcal{N}(0, 1)$ directly into an image latent without any iterative denoising steps, hence $z_0 = \sdturbo (z_T,T,\textembedding)$. In this paper, we build on the state-of-the-art one-step diffusion model SD-Turbo~\cite{sauer2023adversarial} as the backbone for real image inversion and editing.

\subsubsection{DDIM inversion} 
Different from inferring a real image from Gaussian noise as in DDIM sampling, DDIM Inversion aims finding an initial noise $z_T$ that reconstructs the input latent code $z_0$ upon sampling. Since we aim at precisely reconstructing a given image for future editing, we employ the deterministic DDIM sampling~\cite{song2021ddim}:
\begin{equation}
    z_{t+1} = \sqrt{\bar{\alpha}_{t+1}}f_\theta(z_t,t,\textembedding) + \sqrt{1-\bar{\alpha}_{t+1}} \model(z_t,t,\textembedding)
    \label{eq:ddim}
\end{equation}
where $\bar{\alpha}_{t+1}$ is the  scaling factor defined and $f_\theta(z_t,t,\textembedding)$ predicts the final denoised latent  $z_0$ as:
\begin{equation}
    f_\theta(z_t,t,\textembedding) = \Big[z_t - \sqrt{1-\bar{\alpha}_t} \model(z_t,t,\textembedding) \Big] / {\sqrt{\bar{\alpha}_t}}
\end{equation}
We denote the DDIM inversion to return the initial latent $\tilde{z}_T$ by $\tau$ steps as $\tilde{z}_T=DInv_{\tau}$.

\subsection{Training-Free Inversion (\ourmethod)} \label{tfi}

Our method framework diagram is shown in Fig~\ref{fig:method}. Given the real image $\inputimage$ for editing, the first step is to compute an initial noise $\tilde{z}_T = DInv_{\tau}$ through DDIM inversion as the starting point for both reconstruction and editing tasks. 
However, our analysis (Fig.~\ref{fig:anal_noise}) reveals that the misalignment between the image and the text prompt, namely the \textit{caption gap}, significantly impedes convergence during image reconstruction. 
Furthermore, as shown in Fig.~\ref{fig:anal_steps}, naively applying one-step DDIM inversion results in reduced editability.
We hypothesize that this reduction in editability stems from the source image patterns being retained in the inverted initial noise $\tilde{z}_T$. This retention imposes constraints on the \textit{initial latent editability}, a phenomenon consistent with previous observations~\cite{nguyen2024swiftedit,mokady2022null}.
From the view of diffusion model training, the issues of distribution gaps~\cite{zhang2024real_bias} and signal leakage~\cite{everaert2024_signal_leak_bias} naturally arise due to inherent properties of the diffusion process.
These terms describe a shared phenomenon: after DDIM inversion, the latent $z_T$—generated by adding Gaussian noise to $z_0$—deviates from a true standard Gaussian distribution. 
Instead, it retains residual patterns or priors from the input image. This leads to a biased noise distribution during the forward diffusion, causing structural layout to persist even at high noise levels. Such residual bias disrupts the alignment needed for accurate reconstruction in the reverse T2I generation from $z_T$ to $z_0$.
Moreover, this issue becomes more pronounced with fewer inversion steps $\tau$ in the DDIM process. A reduced $\tau$ exacerbates the distribution gap, thereby amplifying the misalignment during reconstruction and further diminishing the flexibility required for effective editing. These findings highlight the critical need for alternative techniques like \ourina and \oursuffix to address these challenges.

\subsubsection{Stage 1: Iterative Noise Alignment (\ourina)} 
After obtaining the initial noisy latent $\tilde{z}_T = DInv_{\tau}$, we reconstruct the predicted real image using the one-step diffusion model $\sdturbo$ as: $\tilde{z}_0 = \sdturbo (\tilde{z}_T,T,\textembedding)$.
Unlike methods that train new T2I diffusion models~\cite{starodubcev2024invertible_cd} or introduce additional modules~\cite{wu2025turboedit_adobe,nguyen2024swiftedit} to bridge the image appearance gap, we iteratively update the latent $\tilde{z}_T$ by minimizing a combination of the reconstruction loss and the KL divergence:

\begin{align}
    \mathcal{L}_{1} & = {\Vert z_0 - \tilde{z}_0 \Vert_{2}^{2}}  + 
    {\lambda \cdot  D_{KL} (\tilde{z}_T || \mathcal{N}(0, 1) )} \\
    \tilde{z}_T & = \tilde{z}_T - \nabla_{\tilde{z}_T} {\mathcal{L}_{1}}
\label{eq:iterna_loss}
\end{align}
where $\lambda$ is a trade-off parameter. Updated with both losses $\mathcal{L}_{mse}$ and $\mathcal{L}_{KL}$, the noisy latent is able to reconstruct the input image $\inputimage$ while getting close to the ideal normal distribution. After updating the latent with $\mathcal{N}_{iter}$ steps, we denote the final noisy latent as $\bar{z}_T = \tilde{z}_T^{\mathcal{N}_{iter}}$ and freeze it in the following processes.

\subsubsection{Stage 2: Suffix Learning (\oursuffix)} 
In the second stage, we aim to enhance the representation of input caption conditions. Instead of directly updating the input prompt embeddings $\textembedding$, which may lead to overfitting to the input image and degrade editability, we introduce several suffix tokens to aid in restoring the input images.
These suffix tokens, denoted as $\langle s^*\rangle$, are randomly initialized and appended to the input text prompt. They serve solely for reconstruction purposes and do not carry any semantic meaning. We further denote the number of suffix tokens as $\mathcal{N}_{s*}$.
For example, for $\mathcal{N}_{s*}=3$, given the input text prompt $\textprompt$=\quotes{A photo of a round cake}, the rewritten prompt with suffix tokens becomes: $\textprompt^*$=\quotes{A photo of a round cake $\langle s1 \rangle \langle s2\rangle \langle s3 \rangle$}. 

The corresponding textual embedding is then updated as: $\textembedding^*=\conditioner(\textprompt^*)$.
With the updated textual embedding, the generated image is obtained as: $\bar{z}_0 = \sdturbo (\bar{z}_T,T,\textembedding^*)$.
During training, the suffix tokens are optimized using the MSE loss:

\begin{align}
    \mathcal{L}_{2} & = {\Vert z_0 - \bar{z}_0 \Vert_{2}^{2}} \\
    s* & = {s*} - \nabla_{s*} {\mathcal{L}_{2}}
\label{eq:suffl_loss}
\end{align}

This ensures the suffix tokens complement the reconstruction process without compromising editing flexibility.

\subsection{Text-Guided Image Editing} \label{tfi_edit}
\label{sec:image_editing}
After performing the proposed diffusion inversion (\ourmethod), we can achieve text-guided image editing by simply modifying the source prompt into the desired target prompt. The naive editing result can be expressed as: $\bar{z}_0^{edit}=\sdturbo (\bar{z}_T,T,\textembedding^{edit})$, where the target prompt embedding is computed as $\textembedding^{edit}=\conditioner(\textprompt^{edit})$. And the editing prompt is also appended with the additional $\mathcal{N}_{s*}$ suffixes $\mathcal{N}_{s*}$ for consistency. For instance, if the user wants to modify “a round cake” into “a square cake”, the editing prompt becomes: $\textprompt^{edit}=$\quotes{A photo of a square cake $\langle s1 \rangle \langle s2\rangle \langle s3 \rangle$} by appending these three new learned suffix tokens.
This simple modification in the prompt allows for flexible and efficient text-guided image editing, leveraging the reconstructed latent and maintaining high editability.

\section{Experiments}

In this section, we provide comprehensive information about the dataset benchmarks and evaluation metrics employed in Section~\ref{Dataset and Evaluation Metrics}. 
Subsequently, we delve into the compared methods in Section~\ref{Compared Methods} and the implementation details in Section~\ref{Implementation Details}. 
Following that, the experimental results, including qualitative and quantitative comparisons, are presented in Sections~\ref{sec:expr_result}. Lastly, we show the ablation study outlined in Section~\ref{Ablation Studies}.

\subsection{Dataset and Evaluation Metrics} \label{Dataset and Evaluation Metrics}
We evaluate the proposed method using the PIE-Bench dataset~\cite{ju2023direct}, which consists of 700 images from both natural and artificial scenes. Each image is annotated with one of 10 editing types. For fair benchmarking, we adopt seven evaluation metrics from Direct-Inversion~\cite{ju2023direct} to assess the quality of image editing.
Specifically, the evaluation includes: 1) Edit prompt-image consistency with both the entire image and the edited regions, measuring the alignment between the edits and the text prompt. 2) Structure-Dist~\cite{tumanyan2022splicing} for assessing structural preservation during editing. 3) Background preservation metrics, including PSNR, LPIPS~\cite{zhang2018lpips}, MSE, and SSIM~\cite{wang2003ssim}, to quantify the retention of unedited regions.
While edit prompt-image consistency evaluates editability, the remaining CLIP-based~\cite{radford2021clip} metrics assess the preservation of the original image’s knowledge and details.
All experiments were conducted on a single NVIDIA A40 GPU, and we report the average performance across all images in the dataset.

\begin{figure*}[t!]
\centering
  \includegraphics[width=0.998\textwidth]{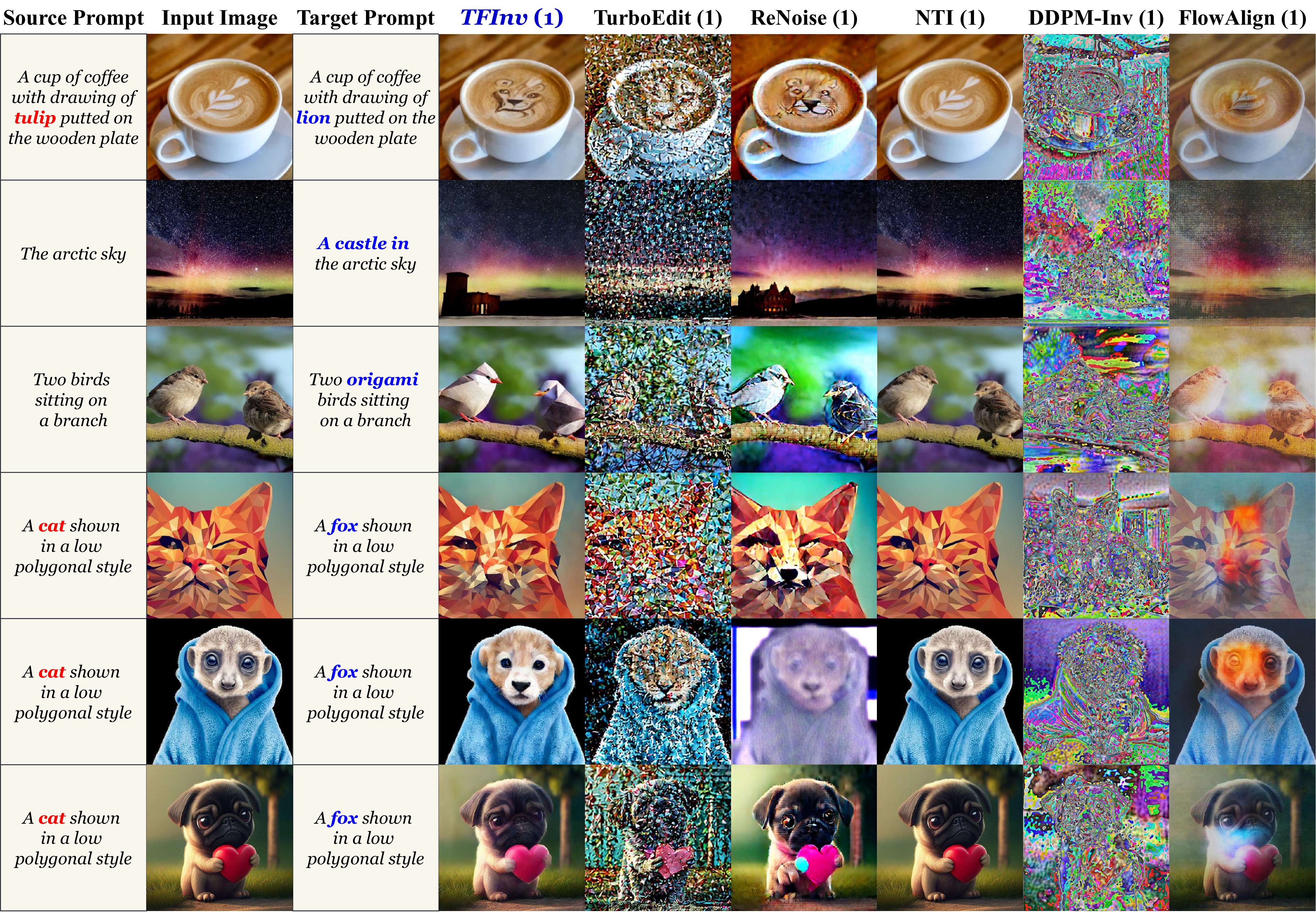}

  \caption{
  Qualitative comparison over the PIE-Bench dataset~\cite{direct_inversion_2023} demonstrates that our method, \ourmethod, surpasses existing approaches in the context of few-step diffusion model-based editing tasks. Numbers in the brackets denote the number of diffusion model steps. 
  }
  \label{fig:editing_compare1}
 \vspace{-4mm}
\end{figure*}


\subsection{Compared Methods} \label{Compared Methods}
We compare our method \ourmethod against several state-of-the-art image inversion and editing approaches designed for few-step diffusion models. Specifically, we include Null-Text Inversion (NTI)~\cite{mokady2022null}, DDPM-Inv~\cite{huberman2024ddpm_friend}, ReNoise~\cite{garibi2024renoise}, TurboEdit~\cite{deutch2024turboedit},
FlowAlign~\cite{kim2025flowalign}
and SwiftEdit~\cite{nguyen2024swiftedit}.
We reproduce their results using the PIE-Bench~\cite{direct_inversion_2023} dataset, a benchmark specifically curated for evaluating text-guided image inversion and editing tasks. 
To ensure a fair comparison, we evaluate these methods under configurations of 1, 2 and 4 sampling steps. This setup highlights the efficiency and effectiveness of each method under constrained diffusion settings, allowing us to benchmark performance on both reconstruction and editing tasks.

\begin{table*}[t!]
\centering
\caption{Quantitative Comparison with NTI~\cite{mokady2022null}, DDPM-Inv~\cite{huberman2024ddpm_friend}, ReNoise~\cite{garibi2024renoise}, TurboEdit~\cite{deutch2024turboedit}, FlowAlign~\cite{kim2025flowalign} and SwiftEdit~\cite{nguyen2024swiftedit} over 7 evaluation metrics derived from the PIE-Bench~\cite{direct_inversion_2023}. 
The one-step NTI is not included and proves ineffective for image editing, as demonstrated in Fig.~\ref{fig:editing_compare1} and Sec.~\ref{sec:expr_result}. Note that SwiftEdit~\cite{nguyen2024swiftedit}method requires training on a large-scale dataset.
}
\resizebox{\linewidth}{!}{
\begin{tabular}{c| c |c |ccc c | cc}
\toprule

\multirow{2}{*}{\diagbox{\textbf{Method}}{\textbf{Metrics}}} & \multirow{2}{*}{\textbf{Steps}} & {\textbf{Structure}}  & \multicolumn{4}{c|}{\textbf{Background Preservation}} & \multicolumn{2}{c}{\textbf{CLIP Similarity}} \\ 
 & & {{\textbf{Distance}$_{^{\times 10^3}}$ $\downarrow$}} & \textbf{PSNR} $\uparrow$     & \textbf{LPIPS}$_{^{\times 10^3}}$ $\downarrow$  & \textbf{MSE}$_{^{\times 10^4}}$ $\downarrow$     & \textbf{SSIM}$_{^{\times 10^2}}$ $\uparrow$    & \textbf{Whole}  $\uparrow$          & \textbf{Edited}  $\uparrow$      \\ 

\midrule
\multirow{2}{*}{NTI+P2P~\cite{mokady2022null}} & 2 & 41.14 & 20.29 & 144.22 & 134.10 & 73.98  & 24.25 & 20.96 \\

 & 4 & 65.46 & 17.79 & 180.01 & 236.21 & 70.77 & 25.97 & 22.57 \\
\midrule
\multirow{3}{*}{DDPM-Inv+P2P~\cite{huberman2024ddpm_friend}} & 1 & 181.14  & 10.44 & 531.64 & 1066.27 & 37.99 & 18.39 & 17.36 \\
 & 2 & 184.58 & 9.98 & 536.54 & 1193.75 & 39.32 & 19.62  & 17.78 \\
 & 4 & 177.51 & 8.92 & 501.01 & 1521.06  & 45.34 & 21.51 & 18.50 \\
\midrule

\multirow{3}{*}{ReNoise~\cite{garibi2024renoise}} & 1 & 263.08 & 9.39 & 559.44 & 1379.86 & 38.71 & 15.64 & 16.49 \\
 & 2 & 267.08 & 9.59 & 571.63 & 1317.77  & 36.91 & 15.60 & 16.44  \\
 & 4 & 274.49 & 9.19 & 581.38 & 1440.29 & 35.90 & 15.64 & 16.41 \\
\midrule
\multirow{3}{*}{TurboEdit~\cite{deutch2024turboedit}} & 1 & 233.07 & 12.44 & 469.09 & 649.82 & 38.24 & 20.76 & 18.82  \\
 & 2 & 99.51 & 16.81 & 266.73 & 264.30 & 59.30 & {25.29} & {22.11}  \\
 & 4 & 79.87 & 16.80 & 227.65 & 271.30 & 64.45 & 24.03 & 21.07 \\

\midrule

\multirow{3}{*}{FlowAlign~\cite{kim2025flowalign}} & 1 & 74.08 & 17.51 & 267.75.09 & 787.48 & 38.24 & 21.56 & 19.55  \\
 & 2 & 72.60 & 17.58 & 185.29 & 263.53 & 79.02 & 21.61 & 19.64  \\
 & 4 & 76.22 & 17.54 & 171.24 & 253.44 & 79.08 & 22.73 & 20.05 \\

\midrule

Swiftedit~\cite{nguyen2024swiftedit} & 1 & 15.76 & 23.33 & 77.00 & 55.00 & 82.14 & 25.16 & 21.25  \\
\midrule
\makecell{\ourmethod (Ours)} & 1 & 19.22 & 24.15 & 79.18 & 56.63 & 82.6 & 24.26 & 21.34  \\
\bottomrule
\end{tabular}
}
\label{tab:quantitative_results}

\end{table*}

\subsection{Implementation Details} \label{Implementation Details}
Our method, \ourmethod, is built upon the SD-Turbo~\cite{sauer2023adversarial} one-step model as the backbone. To optimize both the noisy latent $z_T$ and the suffix tokens $s^*$ during the two updating stages, we employ the AdamW optimizer~\cite{loshchilov2017adamw} with a learning rate of $2\mathrm{e}{-2}$. Two stages totally involve $\mathcal{N}_{iter}=600$ iterations.
The initialization of the noisy latent $\tilde{z}_T$ is achieved through DDIM inversion~\cite{song2021ddim} with $\tau=30$ steps to strike a balance between effective inversion and editability. The trade-off hyperparameter is $\lambda=1.0$ by default.

\subsection{Experimental Results} \label{sec:expr_result}

\subsubsection{Qualitative Comparisons}
The main qualitative comparisons are presented in Fig.~\ref{fig:editing_compare1}. None of the baseline methods achieve satisfactory image editing using one-step diffusion models. Specifically, NTI~\cite{mokady2022null} preserves the input image reconstruction with high fidelity but fails to generate meaningful edits. TurboEdit~\cite{deutch2024turboedit} and DDPM-Inv~\cite{huberman2024ddpm_friend} produce blurry outputs, demonstrating poor performance in both reconstruction and editing. FlowAlign~\cite{kim2025flowalign} is able to respond to the designated editing regions, yet it still fails to generate effective edits.
In general, for few-step diffusion model-based editing, NTI aligns with user target prompts but still struggles to maintain consistency with the input image, particularly distorting the background information. And all the compared methods distort the overall image content during editing, with the background often entirely lost.
In contrast, our proposed method, \ourmethod, achieves significantly better results even with the one-step diffusion model. It demonstrates superior reconstruction quality while faithfully implementing user-intended edits, preserving the integrity of the input image, including its background. This underscores the efficacy of our approach in real-image inversion and editing.

\subsubsection{Quantitative Comparisons}
As corroborated by our qualitative comparisons, the numeric results presented in Table~\ref{tab:quantitative_results} further validate the efficiency and effectiveness of our proposed method, \ourmethod. 
While NTI~\cite{mokady2022null} achieves higher scores in certain reconstruction metrics, it does so at the cost of completely bypassing the image editing stage, resulting in no meaningful edits. This trade-off highlights the limitations of NTI when considering tasks that require both accurate reconstruction and flexible, high-quality editing.
As for the other methods, only TurboEdit~\cite{deutch2024turboedit} shows a marginal advantage in terms of the CLIP-score~\cite{radford2021clip}. However, the qualitative results already decisively demonstrate the ineffectiveness of this method, as it causes significant blurring of the background and a noticeable downgrade in image quality. The large distance to our method in other metrics, such as MSE, LPIPS, and SSIM, further corroborates this observation, confirming that TurboEdit fails to provide a satisfactory solution for image editing when considering both qualitative and quantitative factors.
ReNoise~\cite{garibi2024renoise} and DDPM-Inv~\cite{huberman2024ddpm_friend}, exhibit significantly poorer performance compared to our method \ourmethod. This is evident from both quantitative metrics and qualitative evaluations.

The aforementioned methods are all based on U-Net architectures. We further compare our approach with FlowAlign~\cite{kim2025flowalign}, a flow-based method, which still lags significantly behind our method. In addition, we compare against SwiftEdit~\cite{nguyen2024swiftedit}, a training-based approach, and observe comparable performance.
It is worth noting that SwiftEdit requires large-scale pretraining. In contrast, our method is training-free, relying only on the optimization of tokens and latents, which also results in lower memory consumption.

In general, \ourmethod not only reconstructs images effectively but also enables precise and user-preferred edits, maintaining a balance between editability and reconstruction quality. This demonstrates the robustness and versatility of our approach compared to existing methods.

\begin{table*}[t]
\centering
\caption{Time (in seconds) for 1–50 effective edits across different editing methods.}
\vspace{-3mm}
\begin{tabularx}{0.99\textwidth}{l *{5}{>{\raggedleft\arraybackslash}X}}
\toprule
Method & 1 edit & 5 edits & 10 edits & 20 edits & 50 edits \\
\midrule
NTI+P2P~\cite{mokady2022null} (50 steps) & 134   & 670   & 1,340 & 2,680 & 6,700 \\
DDIM-Inv+P2P~\cite{hertz2022prompt} (50 steps) & 25    & 125   & 250   & 500   & 1,250 \\
ReNoise~\cite{garibi2024renoise} (4 steps) & 5     & 25    & 50    & 100   & 250 \\
 \textbf{\ourmethod (Ours)} (1 step) & 120.4 & 122 & 124 & 128 & 145 \\
\bottomrule
\end{tabularx}
\label{tab:ablation_time}
\end{table*}

\begin{table*}[ht]
\centering
\caption{Ablation study on the cross-attention-based mask mechanism.}
\vspace{-3mm}
\label{tab:ablation_mask}
\begin{tabularx}{0.97\textwidth}{>{\centering\arraybackslash}X >{\centering\arraybackslash}X >{\centering\arraybackslash}X >{\centering\arraybackslash}X >{\centering\arraybackslash}X >{\centering\arraybackslash}X}
\toprule
\multirow{2}{*}{\textbf{Mask}} & \multirow{2}{*}{\shortstack{\textbf{Structure Dist.}\\$_{^{\times 10^3}}\downarrow$}} & \multicolumn{4}{c}{\textbf{Background Preservation}} \\
\cmidrule(lr){3-6}
& & \textbf{PSNR} $\uparrow$ & \textbf{LPIPS}$_{^{\times 10^3}}$ $\downarrow$ & \textbf{MSE}$_{^{\times 10^4}}$ $\downarrow$ & \textbf{SSIM}$_{^{\times 10^2}}$ $\uparrow$ \\
\midrule
\cmark & 8.36 & 28.82 & 41.02 & 24.15 & 86.37 \\
\xmark & 25.04 & 22.76 & 100.48 & 78.12 & 80.76 \\
\bottomrule
\end{tabularx}
\end{table*}

\begin{table*}[t!]
\centering
\caption{Ablation studies on token design and caption sources.}
\vspace{-3mm}
\label{tab:ablation_token_and_config}
\begin{tabularx}{0.97\textwidth}{>{\centering\arraybackslash}X|>{\centering\arraybackslash}X|>{\centering\arraybackslash}X|>{\centering\arraybackslash}X|>{\centering\arraybackslash}X|>{\centering\arraybackslash}X>{\centering\arraybackslash}X}
\toprule
\multirow{2}{*}{\textbf{Setting}} & \multirow{2}{*}{\textbf{Config}} & \multirow{2}{*}{\textbf{\#Tokens}} & \textbf{Structure} & \multirow{2}{*}{\textbf{PSNR $\uparrow$}} & \multicolumn{2}{c}{\textbf{CLIP Similarity}} \\
& & & \textbf{Dist.$_{^{\times 10^3}}$$\downarrow$} & & \textbf{Whole} $\uparrow$ & \textbf{Edited} $\uparrow$ \\
\midrule
\multicolumn{7}{c}{\textit{Token Position \& Number}} \\
\midrule
Prefix & - & 1 & 14.18 & 25.75 & 23.22 & 21.66 \\
Suffix & - & 1 & 25.06 & 23.04 & 24.81 & 22.66 \\
Suffix & - & 3 & 16.36 & 25.52 & 23.82 & 22.02 \\
Suffix & - & 5 & \textbf{13.89} & \textbf{25.73} & 23.08 & 21.07 \\
\midrule
\multicolumn{7}{c}{\textit{Caption Source}} \\
\midrule
- & PIE-Bench & 3 & \textbf{16.36} & \textbf{25.52} & 23.82 & 22.02 \\
- & LLaVA & 3 & 19.89 & 23.93 & \textbf{24.23} & \textbf{22.36} \\
\bottomrule
\end{tabularx}
\end{table*}

\begin{table*}[t!]
\centering
\caption{Ablation study on the hyperparameter $\lambda$.}
\vspace{-3mm}
\label{tab:ablation_hyparam}
\begin{tabularx}{0.97\textwidth}{>{\centering\arraybackslash}X|>{\centering\arraybackslash}X|>{\centering\arraybackslash}X|>{\centering\arraybackslash}X|>{\centering\arraybackslash}X>{\centering\arraybackslash}X}
\toprule
\multirow{2}{*}{\textbf{$\lambda$}} &
\multirow{2}{*}{\shortstack{\textbf{Structure Dist.}\\$\times 10^3\downarrow$}} &
\multirow{2}{*}{\textbf{PSNR $\uparrow$}} &
\multirow{2}{*}{\shortstack{\textbf{SSIM}\\$\times 10^2\uparrow$}} &
\multicolumn{2}{c}{\textbf{CLIP Similarity}} \\
& & & & \textbf{Whole} $\uparrow$ & \textbf{Edited} $\uparrow$ \\
\midrule
0.1 & 15.11 & 25.60 & 85.34 & 23.75 & 21.47 \\
1   & 15.87 & 25.52 & 85.30 & 23.82 & 22.02 \\
5   & 17.17 & 25.12 & 84.95 & 23.76 & 22.10 \\
10  & 17.13 & 25.25 & 84.65 & 23.68 & 21.56 \\
50  & 16.73 & 24.78 & 84.07 & 23.51 & 21.59 \\
\bottomrule
\end{tabularx}
\end{table*}

\subsubsection{Computational Efficiency}

In terms of computational cost, each image undergoes the inversion process only once, which takes approximately 2 minutes. Afterward, each image editing operation requires only a single step and can be completed in about 0.4 seconds. We measured the time required for 1–50 edits and compared it with standard multi-step editing methods with the results presented in Table~\ref{tab:ablation_time}. The results indicate that a single edit takes 120.4 seconds, which is longer than conventional multi-step methods. However, for 50 edits, the total time is only 145 seconds—significantly less than the 6,700 seconds required by NTI+P2P~\cite{mokady2022null} and the 1,250 seconds required by DDIM+P2P~\cite{hertz2022prompt}.

\subsection{Ablation Studies} \label{Ablation Studies}

\subsubsection{Cross-Attention Guided Editing}
\begin{figure*}[t!]
\centering
\begin{minipage}[h]{0.9\linewidth}
    \centering
    \includegraphics[width=\linewidth]{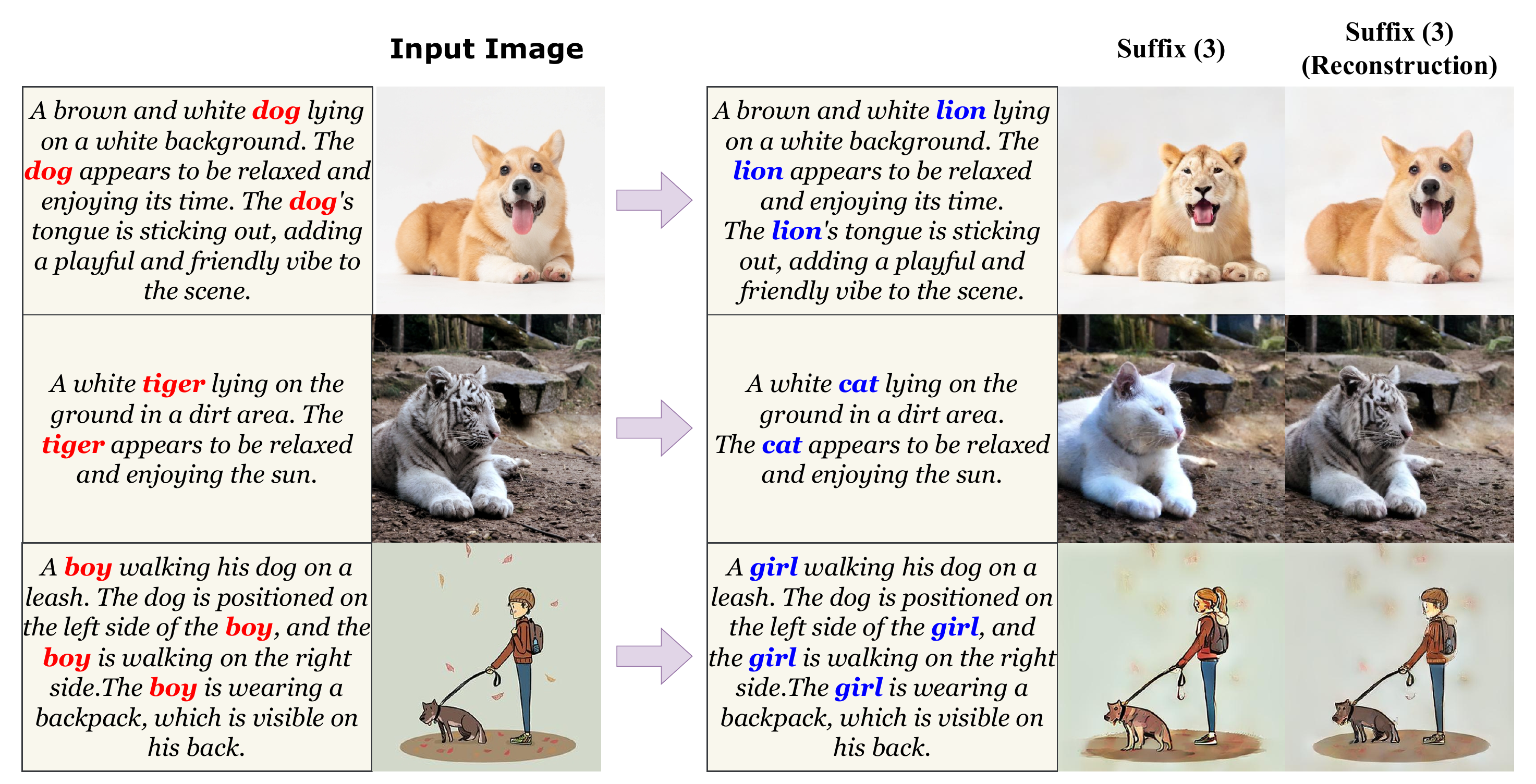}
    \caption{
We further demonstrate the robustness of our method in handling text-guided image editing with longer captions with editing (left) and reconstruction (right) results. Notably, our approach \ourmethod remains effective even when object-related terms are repeated multiple times within the input caption. This highlights its capability to accurately interpret and apply complex textual guidance, ensuring that edits align with the input prompt while preserving the semantic coherence and structural integrity of the images.
}
    \label{fig:ablation_caption}
    
\end{minipage}
\hfill
\begin{minipage}[h]{0.9\linewidth}
    \centering
    \includegraphics[width=\linewidth]{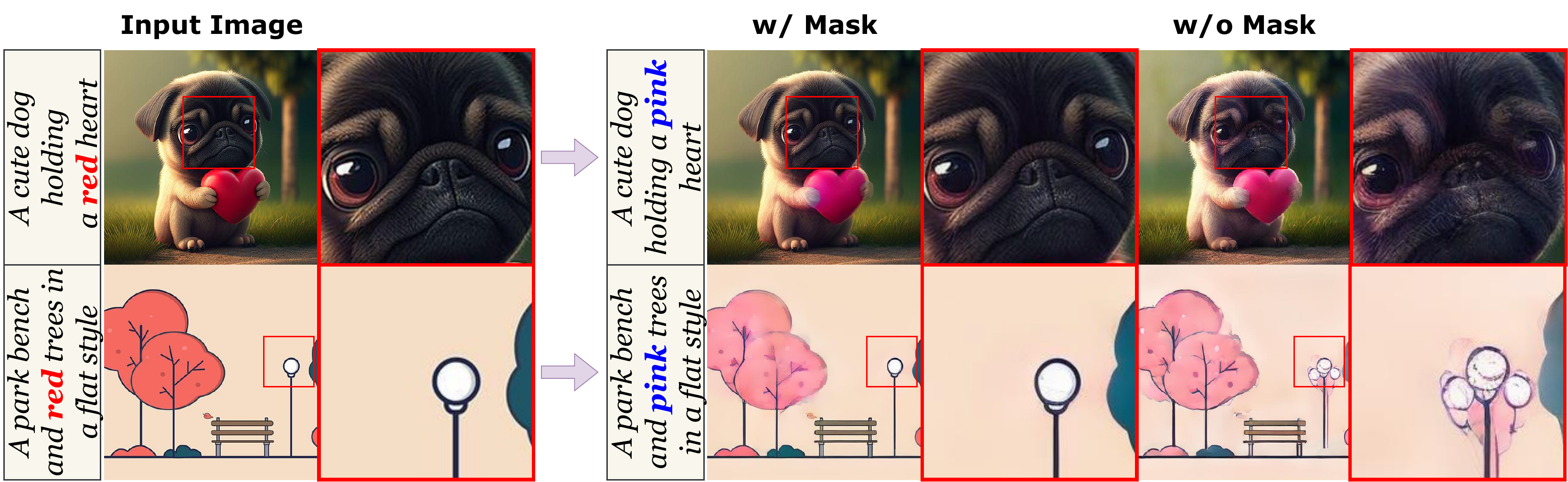}
    \caption{  
  We explore the cross-attention-based mask mechanism and observe that it significantly improves background preservation during the editing process. This advantage is particularly evident in the bottom image, where the street lamp—a critical background element—is well-preserved. 
  }
    \label{fig:ablate_mask}
\end{minipage}

\end{figure*}

For improved background preservation during text-guided image editing, we investigate the localization information encoded within cross-attention maps obtained during the DDIM inversion process. As stated in previous research~\cite{hertz2022prompt,tumanyan2022plug_pnp,kai2023DPL}, these maps inherently contain rich spatial and semantic information about how the model interprets the image regions in relation to the input text prompt.
By visualizing the average cross-attention maps $[\mathcal{A}_t]_{ij}, t \in [1,T]$ during the inversion $DInv_{\tau}$, we derive the average cross-attention maps corresponding to the main object as $[\mathcal{A}]_{obj}$. 
Afterwards, we normalize it into the range $[0,1]$ as mask $\mathcal{M}= [\mathcal{A}]_{obj} / \{ \max ([\mathcal{A}]_{obj}) - \min ([\mathcal{A}]_{obj}) \}$. 
During the editing process, we use the mask to merge the output from the original real image branch and the editing branch as:
\begin{align}
    \hat{z}_0^{edit} = \bar{z}_0^{edit} \cdot \mathcal{M} + \bar{z}_0 \cdot (1-\mathcal{M})
\label{eq:mask_edit}
\end{align}
This formula effectively achieves image editing tasks while ensuring robust background preservation. By leveraging cross-attention maps to guide the fusion of edited and original image latents, it strikes a trade-off between maintaining the background fidelity and applying precise text-guided edits to the target regions. In Fig.~\ref{fig:ablate_mask}, we present the editing effects with and without masks.

Table~\ref{tab:ablation_mask} compares the performance with and without the proposed mask mechanism. Incorporating the cross-attention-based mask mechanism enhances background preservation, ensuring that non-edited regions retain their original structure and appearance. However, this comes at the cost of a slight decrease in image editing performance, particularly in terms of edit prompt alignment.
Given this trade-off, we consider the mask mechanism as an optional choice tailored for tasks that prioritize background integrity in real image editing scenarios.

\subsubsection{Ablate prefix and suffix tokens}
\begin{figure*}[t!]
\centering
  \includegraphics[width=0.999\textwidth]{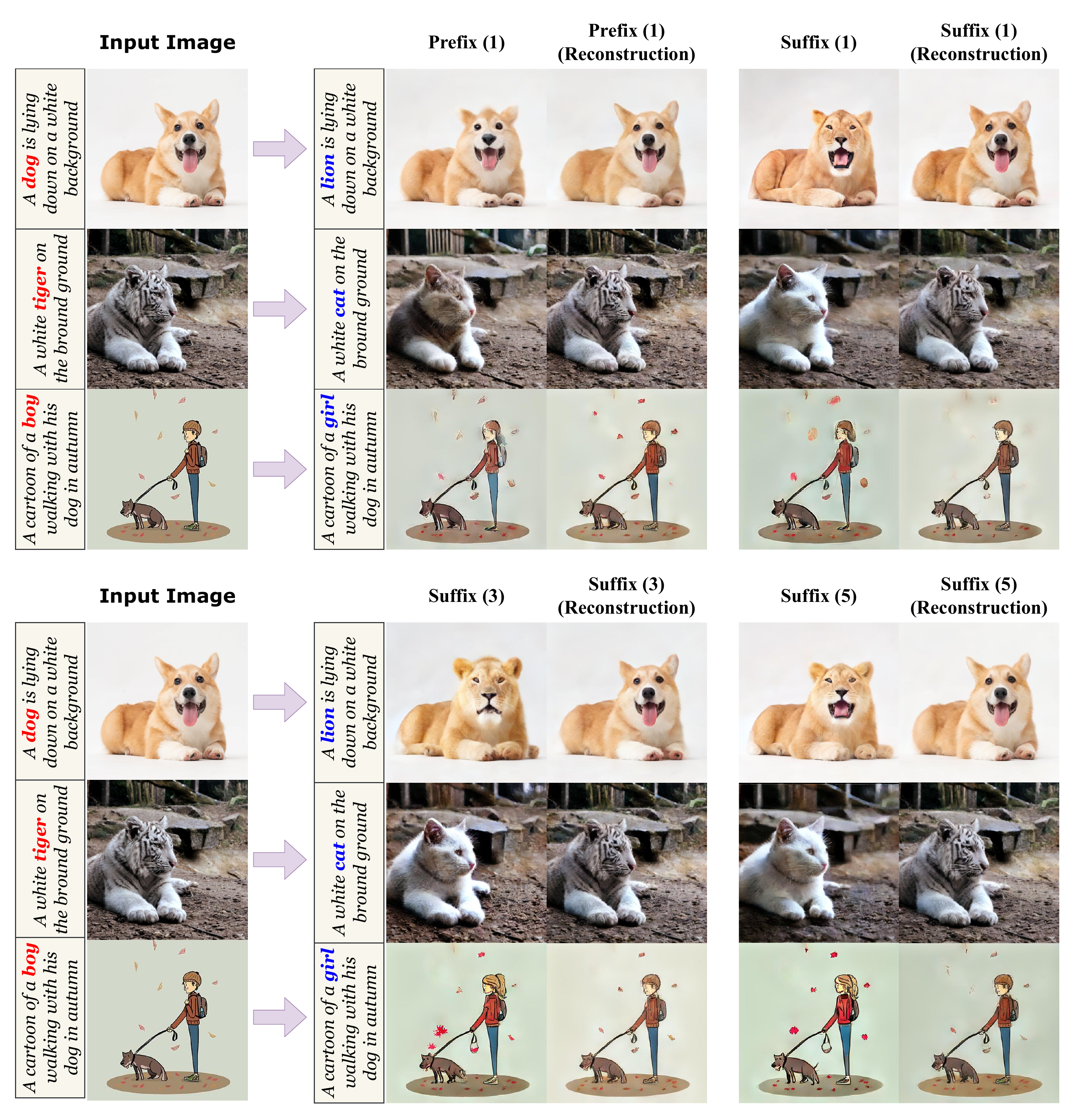}
  \caption{
We analyze the impact of prefix token learning and suffix token learning on performance, considering the number of additional tokens (in brackets) in each case. While prefix tokens contribute positively to image reconstruction (right-side figure), they heavily influence the text embeddings. This strong influence often distorts the alignment between the prompt and the image, thereby compromising editing flexibility. Adding suffix tokens—especially with 3 additional tokens—strikes a balance between reconstruction quality and editing capability. The suffix tokens operate independently of the main prompt semantics, relieving the caption gap while preserving the inherent flexibility required for effective image editing.
  }
  \label{fig:tokens_num}
\end{figure*}

Table~\ref{tab:ablation_token_and_config} and Fig.\ref{fig:tokens_num} present the ablation results on the usage of prefix and suffix tokens, which aims to solve the \textit{caption gap} issue. While applying prefix tokens can achieve reasonable image reconstruction, it significantly compromises the editing capability due to the substantial distortion introduced to the text embeddings.
In contrast, leveraging suffix token learning ensures both accurate reconstruction and effective editing. Notably, the use of 3 suffix tokens strikes an optimal trade-off, balancing the fidelity of reconstruction and the flexibility of editing in our proposed method, \ourmethod.

\subsubsection{Caption Source}
Table~\ref{tab:ablation_token_and_config} provides an analysis of the influence of caption sources on both inversion and editing performance. Among the evaluated sources, LLaVa~\cite{touvron2023llama}-generated captions demonstrate superior ability to describe input images accurately, enhancing text-image alignment during editing. 
While the improved alignment contributes to better editing outcomes, it comes at a slight cost in terms of preserving the original image’s structure and background integrity. These findings underscore the trade-off between semantic alignment and structural fidelity, emphasizing the importance of carefully selecting or generating captions tailored to specific editing objectives.

Additionally, Fig~\ref{fig:ablation_caption} demonstrates that \ourmethod is robust in text-guided image editing with longer captions. It effectively handles repeated object-related terms, accurately applying complex textual guidance while preserving the semantic coherence and structural integrity of the images.

\subsubsection{Hyperparameters}
We conducted an ablation study to systematically determine appropriate hyperparameters for $\lambda$ in Eq.~\ref{eq:iterna_loss}. We varied  $\lambda$ across a range of [0.1, 50.0] and evaluated the impact on both image reconstruction and editing performance. The results of five representative trials are summarized in Table~\ref{tab:ablation_hyparam}.
Notably, our experiments indicate that  $\lambda$ is the optimal choice, achieving the best balance between maintaining input image fidelity during reconstruction and enabling effective user-guided editing. As such,  $\lambda=1.0$ has been set as the default value throughout this paper. These findings highlight the sensitivity of $\lambda$ and its critical role in balancing reconstruction and editing performance.

\subsubsection{Backbone}

To further evaluate the effectiveness and generalizability of the proposed method, we conduct experiments across different generative backbones. In addition to the results on SD-Turbo, we apply \ourmethod to the LCM model for validation.

\begin{figure}[ht]
\centering
  \includegraphics[width=0.49\textwidth]{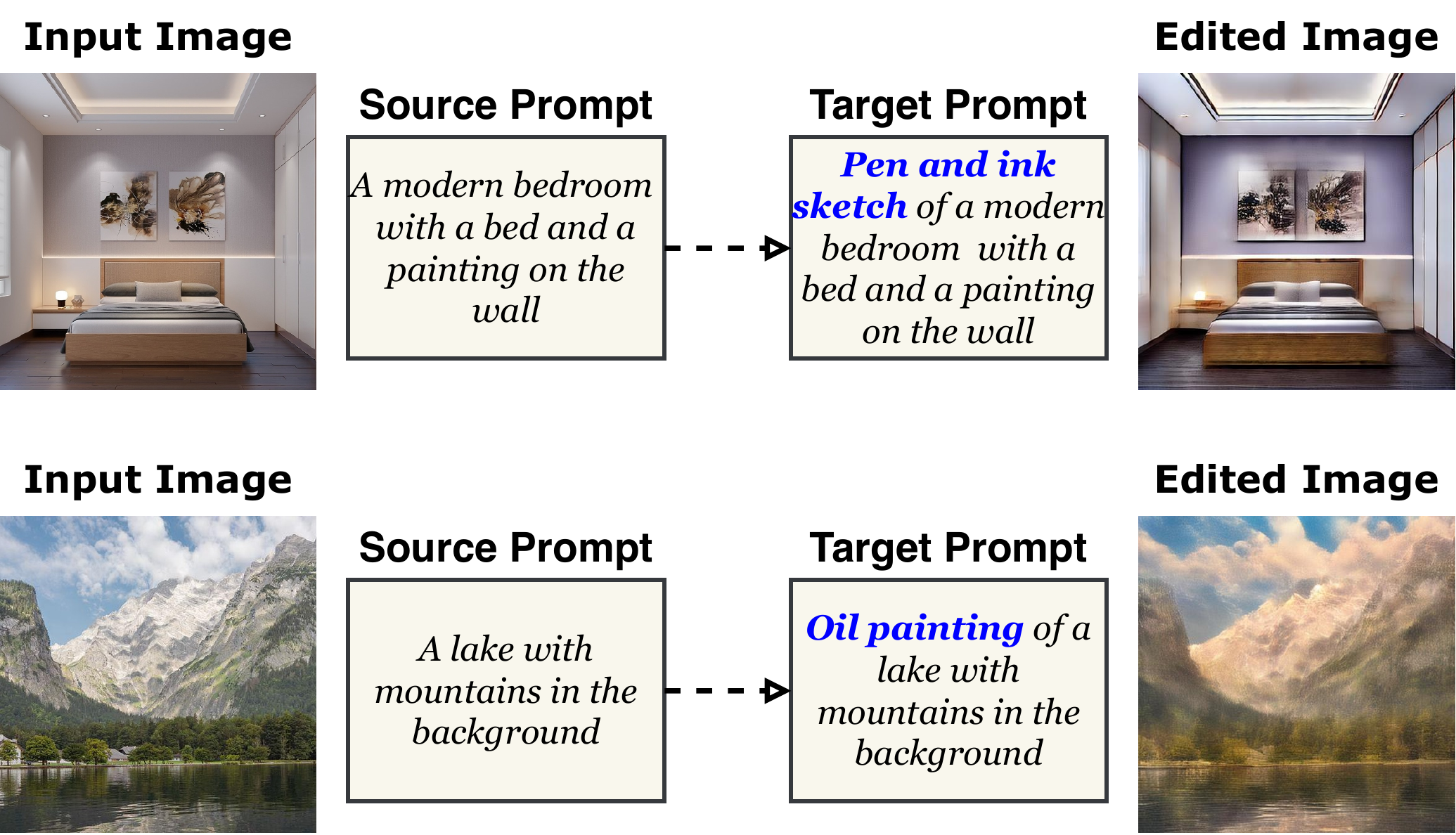}
  \caption{ 
    Qualitative comparison of editing results across different generative frameworks (SD-Turbo and LCM), demonstrating the generalization capability of the proposed method.
  }
  \label{fig:ablation_backbone}
\end{figure}

As shown in Fig.~\ref{fig:ablation_backbone}, even when the underlying generative model is replaced, our method consistently achieves reliable and accurate editing results, while preserving semantic consistency and structural details. These results demonstrate that the proposed approach is not tied to a specific generative framework and exhibits strong generalization capability.

\subsection{Limitations and Discussions}

\subsubsection{Limitations}

The proposed method has several limitations. While the inference process requires only one step for editing, the inversion procedure remains computationally intensive: a standard setting of 600 optimization steps takes approximately 2 minutes per image. Although inversion is required only once per image, the overall runtime is still dominated by this stage, which limits scalability in large-scale or real-time applications.This observation reflects a trade-off between training-free editing flexibility and computational efficiency. In addition, the number of optimization steps is also a factor to consider. While too many optimization steps can preserve the background well, they can weaken editing capabilities, which reflects the trade-off between reconstruction and editing.

\subsubsection{Discussions}

The proposed method enables image editing without requiring model fine-tuning, making it readily applicable to pretrained diffusion models. This advantage significantly lowers the barrier to deployment and removes the dependence on additional training data or costly retraining procedures. Such flexibility highlights the potential of the method for practical and scalable applications.

However, the current approach still involves a relatively high optimization cost, which may limit its efficiency in real-world scenarios. Future work could address this limitation by investigating more effective initialization strategies and designing more efficient optimization algorithms, thereby improving both computational efficiency and usability.

In addition, the method is currently restricted to U-Net–based architectures. Considering the growing importance of transformer-based diffusion frameworks, such as MMDiT, extending the proposed approach to these models represents an important direction for future research and could further broaden its applicability.

\section{Conclusion}
In this paper, we introduce a novel, training-free inversion (\ourmethod) framework for text-guided image editing utilizing one-step diffusion models,
overcoming the caption gap and initial latent editability issues.
By proposing innovative techniques such as iterative noise alignment (\ourina) and suffix learning (\oursuffix), we enable accurate real image inversion and seamless editing within the one-step diffusion framework. Furthermore, our cross-attention-based mask editing mechanism ensures precise, localized modifications while preserving background details, offering a unique balance between flexibility and efficiency. Extensive experiments demonstrate that our method, \ourmethod, achieves state-of-the-art editing performance, outperforming existing methods in terms of both speed and quality. This work represents a significant step toward making high-quality, real-time image editing practical for a wide range of applications, including on-device deployments, paving the way for future advancements in the field.

\bibliographystyle{unsrtnat}
\bibliography{longstrings,mybib}





\end{document}